\RequirePackage{amsmath}
\documentclass[runningheads]{llncs}
\usepackage[T1]{fontenc} 
\usepackage[table]{xcolor}
\usepackage{graphicx}
\usepackage{times} 
\usepackage{amsmath} 
\usepackage{amssymb}  
\usepackage[T1]{fontenc}
\usepackage{graphicx}
\usepackage{multirow}
\usepackage{tabularx,booktabs}
\usepackage{subcaption}
\usepackage[inkscapelatex=false]{svg}
\usepackage[flushleft]{threeparttable}
\usepackage{algorithm}
\usepackage{algpseudocode}
\usepackage{wrapfig}
\usepackage{tabularx}
\usepackage{hyperref}

\usepackage{booktabs}
\usepackage{array}
\begin{document}
%
\title{UAV-Enhanced Combination to Application: Comprehensive Analysis and Benchmarking of a Human Detection Dataset for Disaster Scenarios}
\titlerunning{UAV-Enhanced Combination to Application}
%
\author{Ragib Amin Nihal\inst{1} \and Benjamin Yen\inst{1} \and Katsutoshi Itoyama\inst{2} \and Kazuhiro Nakadai\inst{1}}
\authorrunning{Nihal et al.}
%
\institute{Tokyo Institute of Technology \and
Honda Research Institute Japan Co.Ltd \\
\email{\{ragib, benjamin, itoyama, nakadai\}@ra.sc.e.titech.ac.jp}}
\maketitle              
\vspace{-.7cm}
\begin{abstract}
Unmanned aerial vehicles (UAVs) have revolutionized search and rescue (SAR) operations, but the lack of specialized human detection datasets for training machine learning models poses a significant challenge.
To address this gap, this paper introduces the Combination to Application (C2A) dataset, synthesized by overlaying human poses onto UAV-captured disaster scenes. Through extensive experimentation with state-of-the-art detection models, we demonstrate that models fine-tuned on the C2A dataset exhibit substantial performance improvements compared to those pre-trained on generic aerial datasets. \textcolor{black}{Furthermore, we highlight the importance of combining the C2A dataset with general human datasets to achieve optimal performance and generalization across various scenarios.} This points out the crucial need for a tailored dataset to enhance the effectiveness of SAR operations. Our contributions also include developing dataset creation pipeline and integrating diverse human poses and disaster scenes information to assess the severity of disaster scenarios. Our findings advocate for future developments, to ensure that SAR operations benefit from the most realistic and effective AI-assisted interventions possible. The dataset, code, and model are publicly available at: \url{https://github.com/Ragib-Amin-Nihal/C2A}

\end{abstract}
\section{Introduction}
The advancement of UAVs, colloquially known as drones, has signaled a new era in the field of emergency response and disaster management. With their unparalleled agility and ability to provide an aerial perspective, drones have rapidly become indispensable assets in the arsenal of SAR operations worldwide. These technological marvels significantly improve the efficiency and effectiveness of missions aimed at locating and aiding people in disaster-hit areas \cite{kucukayan2024yolo}. Drones can have a significant impact on minimizing the aftermath of disasters through time efficiency, making a difference in survival and fatality rates.
\\
Despite these advancements, a major shortcoming exists in the deployment of drone technologies—particularly in the area of object detection via drone vision. Existing computer vision or drone vision systems significantly depend on datasets to train detection algorithms. However, these datasets are primarily designed for general situations and do not adequately address the specialized and intricate requirements of disaster contexts. The shortage of disaster detection datasets is mostly owing to logistical and ethical obstacles in capturing and annotating real events, which need substantial resources and often involve sensitive circumstances. The ethical dilemmas of capturing vulnerable people during actual catastrophes aggravate the issue. This lack of specialized human detection datasets for SAR operations hinders the capability of drones to effectively identify human figures disaster scenarios. 
This deficiency is particularly acute, as our findings indicate current pre-trained detection models fall short of effectively identifying humans (more discussed in Section \ref{model_opti}) amidst the multifaceted chaos of disaster scenes—where the stakes are very high.
\\
\begin{figure}[htbp]
    \centering
    \includegraphics[width=\linewidth]{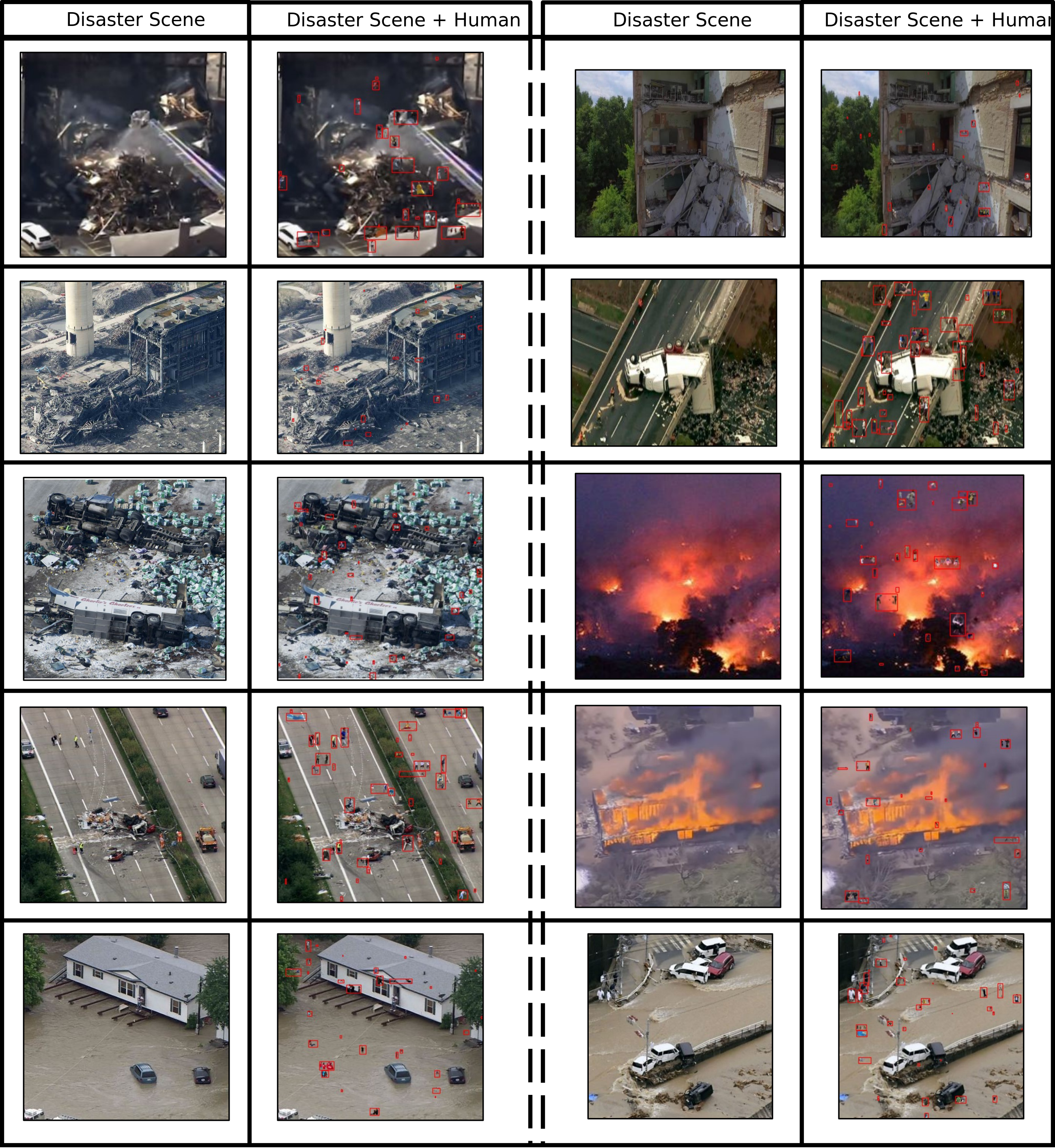}
    \caption{This collection of images presents a selection from our proposed ``Combination to Application" (C2A) dataset, a specialized compilation designed to refine machine learning algorithms for SAR operations in diverse disaster scenarios. Within the bounding boxes, human figures are superimposed onto various disaster scenes, demonstrating the intricate process of overlaying accurately segmented human poses onto different disaster backdrops such as rubble, traffic incidents, flood, and fire. This synthetic approach is crucial for creating more challenging training conditions that AI models may encounter in actual SAR missions. Furthermore, the dataset is enriched with detailed pose information—such as bent, kneeling, lying, sitting, and upright—providing comprehensive data for AI to learn and recognize human forms even when partially occluded by environmental obstacles.}
    \label{fig:enter-label}
\end{figure}
\noindent In critical scenarios, the lack of datasets with the necessary detail for training advanced machine learning algorithms hinders the optimization of drone capabilities. Recognizing the critical need for a specialized dataset, our research introduces a novel dataset explicitly designed to improve human detection capabilities in disaster scenarios through drone vision. This dataset is created by combining human posture images with disaster scene backgrounds, resulting in an intricate collection of images that simulate the diverse settings faced during real-world SAR operations. By  integrating human figures into a variety of disaster scenes, our approach aims to mirror the array of challenges that SAR drones are likely to face, facilitating the development of more robust and effective machine learning models tailored for disaster response applications \cite{nunez2024transformer}.
\\
Our endeavor is inspired by the growing field of disaster management technology, where recent breakthroughs highlight the revolutionary potential of machine learning and computer vision in improving SAR operations \cite{Aryal}. Our research focuses on creating a challenging dataset to train AI models for detecting partially occluded humans, a common scenario in disaster-stricken environments. Such occlusions, resulting from individuals being trapped under debris or obscured by various objects, represent a significant hurdle for human detection systems. Existing datasets scarcely address this challenge, primarily due to the inherent difficulty in replicating these complex scenarios accurately. Our dataset, therefore, signifies a pioneering step towards addressing this gap, offering a resource that simulates the conditions under which SAR operations unfold in reality.
\\
We aim to enhance the precision and reliability of drone-operated human detection in emergency scenarios, underscoring the nuanced requirements of effective disaster response strategies. The construction of this dataset involved a process of
incorporating elements of obstruction to emulate the visibility challenges frequently encountered in real-world disaster settings. This approach facilitates the training of AI models that are not only adept at recognizing human forms in clear view, but are also capable of inferring the presence of individuals in less than optimal visibility conditions. Such abilities are necessary for quickly finding emergency areas and finding survivors who need help right away, which makes SAR missions much more effective  \cite{alsamhi2022uav,hu2022seeing}.
\\
Through this comprehensive paper, we delineate the process undertaken to create our novel synthesized dataset (Fig. \ref{fig:enter-label}), emphasizing its designed complexity and the specific challenges it poses to AI models. We delve into the selection criteria for the images, the sophisticated image processing techniques employed, and the rationale underlying the dataset's structure. We also present our findings from using this dataset to train deep learning models, which show notable improvements in detection accuracy and the feasibility of operational deployment.  In essence, our research essentially provides the field of disaster management with a substantial C2A dataset. Researchers and professionals can use this dataset to extend the potential of drone technology, which will make emergency responses and life-saving operations more effective in chaotic situations. This paper presents our contributions to the field as follows:
\begin{itemize}
    \item[--] We introduce a novel synthesized dataset that mitigates the gap in current SAR operations by providing imagery capturing human figures in disaster contexts, designed to train machine learning models for complex human detection tasks.
    \item[--] We present a comprehensive dataset creation pipeline that combines advanced image processing techniques and domain-specific knowledge, resulting in a dataset that represents the complexity and unpredictability of disaster scenarios. 
    \item[--] Our dataset includes a variety of human poses and disaster scene information, allowing researchers to develop models that can assess the severity of a disaster scene and prioritize rescue efforts.
    \item[--] Our preliminary results demonstrate that the dataset significantly improves the detection performance and operational feasibility of deep learning models, indicating the dataset's potential to transform disaster response efforts.
\end{itemize}
\section{Literature Review}
\label{sect1}
Robust datasets are fundamental to developing and training precise machine learning models. The importance of datasets tailored to specific application domains is well established in the literature. Studies such as those by \cite{alali2024techniques} and \cite{tang2024review} emphasize the need for datasets that encompass the intricacies of various disaster scenarios. These resources are pivotal for calibrating UAV-operated detection systems to recognize human subjects under a multitude of conditions. Yet, there remains a scarcity of datasets that accurately mirror the complexities of disaster-hit environments.
\\
Debris and destruction frequently obscure human subjects in disaster zones, biasing UAV detection. According to the literature, existing datasets and models are progressing, but they fall short of providing the granular detail required for reliable detection in such complex circumstances \cite{munawar2021uavs,malandrino2019planning}. A significant body of research, including work by \cite{kang2023residential}, emphasizes the pressing need for advancing UAV technology to navigate these obstacles adeptly. Yet, the development of datasets that reflect the reality of partial occlusions in disaster contexts is still in its early stages, indicating a pivotal area for future research.
\\
The evolution of machine learning and computer vision has been enhancing to UAV capabilities. These advancements have paved the way for more nuanced data analysis, crucial for discerning human presence within complex terrains. Even with these improvements, the research shows that algorithms and models still need to be improved, especially to make human detection more reliable \cite{khial4725375online}.
\\
\textcolor{black}{One notable effort in this direction is the Search and Rescue Drone (SARD) dataset \cite{sambolek2021automatic}, which focuses on human detection in search and rescue operations using drone imagery. The SARD dataset includes images of people in various poses simulating exhausted or injured individuals, captured in non-urban environments. While it provides a valuable resource for developing detection models for SAR scenarios, the SARD dataset does not explicitly include disaster scenes, which pose additional challenges such as debris, occlusions, and clutter. Moreover, although the dataset contains diverse human poses, this information is not directly incorporated into the annotations, limiting its utility for pose-aware detection.}
\\
The main gap is the absence of comprehensive datasets capturing the full spectrum of disaster scenarios, specifically focusing on the aspect of partial human occlusion. Moreover, there is an evident need for further exploration into advanced machine learning and computer vision applications tailored for SAR operations.
The new dataset is compared to existing datasets about finding people in disaster situations in Table \ref{tab:dataset_comparison}.

\begin{table}[t]
\centering
\caption{Comparison of Datasets for Human Detection in Disaster Scenarios}
\label{tab:dataset_comparison}
\begin{tabular}{|m{1.5cm}|m{2cm}|m{2cm}|m{2cm}|m{2cm}|m{2cm}|}
\hline
\textbf{Feature} & \textbf{C2A (proposed)} & \textbf{LoveNAS \cite{wang2024lovenas}} & \textbf{Smoke Scene \cite{wu2023dataset}} & \textbf{M4SFWD \cite{wang2024m4sfwd}} & \textcolor{black}{\textbf{SARD} \cite{sambolek2021automatic}} \\ \hline
\textbf{Focus} & Human detection in disaster scenarios with partial occlusion & Land-cover mapping for multiple scenes including disaster scenarios & Detection of smoke scenes from satellite imagery for early disaster response & Synthetic dataset for remote sensing forest wildfires detection & \textcolor{black}{Human detection in search and rescue operations using drone imagery} \\ \hline
\textbf{Scenarios Covered} & Various disaster scenarios (earthquakes, flood, fire) with partial occlusion & Urban, rural, and disaster scenes & Areas prone to fire outbreaks & Forest and wildfire scenarios & \textcolor{black}{Search and rescue operations in non-urban areas (no disaster scene)} \\ \hline
\textbf{Partial Occlusion} & Yes & No & No & No & \textcolor{black}{Yes} \\ \hline
\textbf{Image Diversity} & High, with images from multiple disaster types including occluded humans & High, including three normal and two disaster scenes & Moderate, focused on smoke detection & High, developed through post-processing and synthesis & \textcolor{black}{Moderate, focused on search and rescue scenarios} \\ \hline
\textbf{Realism} & Moderate, designed to mimic real disaster conditions with individuals & Moderate, lacks specific focus on human detection & High for smoke detection, moderate for overall disaster realism & High for wildfires, moderate for human detection relevance & \textcolor{black}{High, includes realistic search and rescue scenes captured by drones} \\ \hline
\textbf{Human Poses} & Bent, Kneeling, Lying, Sitting, Upright & Not applicable & Not applicable & Not applicable & \textcolor{black}{Diverse, not included in annotation} \\ \hline
\end{tabular}
\end{table}

\noindent This comparison indicates the distinctive contributions of the newly developed dataset, particularly its emphasis on effective human detection in disaster scenarios, a topic not explicitly addressed by the existing datasets. The newly developed dataset sets itself apart from existing ones, which narrowly focus on specific types of disaster scenes or aspects like smoke and fire detection.
\\
The new dataset mitigates in a gap in specifically designed datasets for finding partially occluded individuals in disaster scenarios. Its goal is to greatly enhance the abilities of machine learning models for disaster response and SAR operations, ultimately leading to more effective and timely humanitarian efforts.
\section{Dataset Creation Pipeline}
\label{sect2}
We developed a systematic pipeline to produce a comprehensive set of images for training machine learning models to detect humans in disaster scenarios. The dataset combines parts of the Aerial Image Dataset for Emergency Response Applications (AIDER) and the LSP/MPII-MPHB dataset. It shows a variety of human poses on a range of disaster backgrounds.
\subsection{Data Sources and Composition}
\textbf{AIDER (Aerial Image Dataset for Emergency Response Applications):} The AIDER dataset \cite{kyrkou2019deep} serves as the foundation for the disaster scene backgrounds. It comprises images from four major disaster types: Fire/Smoke (320 images), Flood (370 images), Collapsed Building/Rubble (320 images), and Traffic Accidents (335 images). These authentic disaster images offer a realistic portrayal of the chaotic and unpredictable conditions typical in emergency scenarios.  We did not utilize the 1,200 normal case images to keep the focus on emergency situations. This dataset offers a glimpse into the chaotic and unpredictable environments that characterize disaster scenes, making it an ideal choice for our purposes.\\ \\
\textbf{LSP/MPII-MPHB (Multiple Poses Human Body):} For the human subjects, we sourced images from the LSP/MPII-MPHB dataset \cite{andriluka20142d,johnson2010clustered}, which contains 26,675 images featuring 29,732 instances of human bodies in various poses. This dataset is specifically designed to capture a wide range of human body positions, including bent, kneeling, sitting, upright, and lying, providing the necessary diversity to train models for detecting humans under different conditions. The detailed annotations of human poses in this dataset are critical for training models to recognize human figures in complex disaster environments.
\vspace{-.5cm}
\subsection{Pipeline Steps}
\begin{algorithm}[htbp]
\caption{Dataset Creation Pipeline for Human Detection in Disaster Scenarios. This pipeline integrates human poses with disaster scene backgrounds, involving background removal, cropping, random scaling, and overlaying to simulate realistic disaster environments for training AI models.}
\label{algorithm}
\begin{algorithmic}[1]

\State \textbf{Input:} 
\State $AIDER$: Set of disaster scene images
\State $MPHB$: Set of human pose images
\State \textbf{Output:} Combined dataset $D$ with annotated human poses in disaster scenes

\Procedure{U2NetRemoveBackground}{$image$}
    \State Apply U2-Net model to $image$ for segmentation
    \State Extract foreground (human) based on segmentation result
    \State \textbf{return} foreground
\EndProcedure

\Procedure{CropFocusedObject}{$image$}
    \State Compute bounding box around non-zero pixels in $image$
    \State Crop $image$ to the bounding box
    \State \textbf{return} cropped image
\EndProcedure

\Procedure{RandomScale}{$image$}
    \State $scale \gets$ random value between predefined min and max
    \State Resize $image$ by $scale$
    \State \textbf{return} resized image
\EndProcedure

\Procedure{RandomPosition}{$background, object$}
    \State $bgWidth, bgHeight \gets$ dimensions of $background$
    \State $objWidth, objHeight \gets$ dimensions of $object$
    \State $x \gets$ random integer from $0$ to $bgWidth - objWidth$
    \State $y \gets$ random integer from $0$ to $bgHeight - objHeight$
    \State \textbf{return} $(x, y)$
\EndProcedure

\Procedure{CreateDataset}{$AIDER, MPHB$}
    \State $D \gets \emptyset$
    \For{each pose $p$ in $\{bent, kneeling, sitting, upright, lying\}$}
        \For{each image $i$ in $MPHB$ corresponding to pose $p$}
            \State $i_{bg\_removed} \gets \Call{U2NetRemoveBackground}{i}$
            \State $i_{cropped} \gets \Call{CropFocusedObject}{i_{bg\_removed}}$
            \If{Size of $i_{cropped}$ $\geq 0.02 \times$ Size of $i_{bg\_removed}$}
                \State Add $i_{cropped}$ to $MPHB_p$
            \EndIf
        \EndFor
    \EndFor
    \For{each image $a$ in $AIDER$}
        \State $H \gets \text{Random selection of human poses from } MPHB_p$
        \For{each human pose $h$ in $H$}
            \State $h_{scaled} \gets \Call{RandomScale}{h}$
            \State $pos \gets \Call{RandomPosition}{a, h_{scaled}}$
            \State Overlay $h_{scaled}$ on $a$ at position $pos$
            \State Compute bounding box $bbox$ for $h_{scaled}$ at $pos$
            \State Add $(a, bbox)$ to $D$
        \EndFor
    \EndFor
    \State \textbf{return} $D$
\EndProcedure

\end{algorithmic}
\end{algorithm}

\subsubsection{1. Background Removal and Image Preparation:}
Using the U2Net segmentation model \cite{qin2020u2}, we isolated human figures from the LSP/MPII-MPHB dataset by removing the background. The U2-Net, short for "U-Squared Net," is a deep neural network known for its powerful performance in salient object detection and image segmentation tasks. It employs a nested U-structure that enhances the learning of local and global features within images, enabling precise segmentation of objects, including human figures, from their backgrounds. This process involved saving each figure with its respective pose in a separate folder, ensuring that the focus remained on the human subject without any background distractions.
\vspace{-.3cm}
\subsubsection{2. Image Cropping and Cleaning:}
In the next step, the isolated images were then cropped to highlight the human figures, removing unnecessary peripheral content. This step involved calculating the minimum and maximum indices of non-zero pixel elements to determine the bounding box for each figure. Images where non-zero indices constituted less than 2\% of the total image area were excluded to minimize noise and inaccuracies.
\vspace{-.3cm}
\subsubsection{3. Overlay Process:}
For each disaster background from the AIDER dataset, human figures from the LSP/MPII-MPHB dataset were overlaid at random positions. The scaling of human figures in the dataset was randomized within specified lower and upper bounds, taking into account the dimensions of the disaster scene backgrounds. This approach was employed to mimic the diverse scales at which humans may be observed in real disaster scenarios. This process also included checks for collisions and adjustments to the placement of figures to ensure a realistic composition. The final images were annotated with bounding boxes, accurately reflecting the position and scale of each human figure within the disaster scene.
\\
The dataset creation pipeline integrates human poses with disaster scene backgrounds to construct a comprehensive dataset for training machine learning models for human detection within disaster scenarios. The Algorithm \ref{algorithm} outlines the steps involved in the dataset creation process.
\section{Properties of C2A Dataset}
\label{sect3}
The C2A (Combination to Application) dataset\footnote[1]{Dataset available at: \url{https://github.com/Ragib-Amin-Nihal/C2A}} is a curated collection specifically designed for advancing human detection disaster scenarios by combining AIDER dataset images (disaster scene backgrounds) and diverse human poses from  the LSP/MPII-MPHB dataset. Some of the samples of the dataset are presented in Fig. \ref{fig:enter-label}. Comparison of various datasets is shown on Table \ref{tab:dataset_comparison2}. In this section, we present a comprehensive analysis of the dataset's properties.
\vspace{-.8cm}
\begin{table}[t]
\centering
\caption{Comparison of various datasets \cite{xia2018dota} including the proposed C2A dataset}
\label{tab:dataset_comparison2}
\begin{tabular}{@{}|l|c|c|c|c|r|@{}}
\toprule
Dataset & Annotation way & \# main categories & \# Instances & \# Images & Image width \\ \midrule
SARD & horizontal BB & 1 & 6,532 & 1,981 & 1920 \\
M4SFWD & oriented BB & 2 & 17,613 & 3,946 & 776--1480 \\
Smoke Scene & oriented BB & 2 & 18,849 & 8,735 & 95--6000 \\
Tiny Persons & horizontal BB & 1 & 70,702 & 1,570 & 765--2048 \\
Crowd Human & horizontal BB & 2 & 456,098 & 19,370 & 400--10800 \\
PASCAL VOC & horizontal BB & 20 & 27,450 & 11,530 & 640 \\
MS COCO & horizontal BB & 80 & $\sim$2.5M & $\sim$328,000 & 640 \\
NWPU VHR-10  & horizontal BB & 10 & 3,651 & 800 & $\sim$1000 \\
3K Vehicle Detection & oriented BB & 2 & 14,235 & 20 & 5616 \\
DOTA & oriented BB & 14 & 188,282 & 2,806 & 800--4000 \\
\textbf{C2A (proposed)} & \textbf{horizontal} & \textbf{1 (with 5 poses)} & \textbf{$>$360,000} & \textbf{10,215} & \textbf{150--3400} \\ 
\bottomrule
\end{tabular}
\end{table}

\vspace{.5cm}
\subsection{Number of Images and Image Size}
In the C2A dataset, the total number of images is $10,215$, encompassing over $360,000$ objects for human detection within disaster scenarios. The original size of the images spans a wide range from approximately \( 123 \times 152 \) pixels to high-resolution images of \( 5184 \times 3456 \) pixels. This range is significantly broader than what is commonly found in standard datasets like PASCAL VOC or MSCOCO, where the image dimensions generally do not exceed \( 1000 \times 1000 \) pixels. The wide range of resolution in the C2A dataset ensures the inclusion of various granular details necessary for the precise detection of humans in diverse and challenging disaster environments. Furthermore, the most common image width range within the C2A dataset is between \( 322 \) and \( 600 \) pixels, with over 50.32\% of images falling within this range. The median image width is noted at \( 428 \) pixels, indicative of the dataset's central tendency toward mid-range resolutions. The dataset preserves the integrity of the scenes and avoids potential complications that may arise from segmenting an instance across multiple image pieces.

\subsection{Objects Size}

In our C2A dataset, the pixel size of objects is distributed across a broad spectrum, accommodating the real-world variability in human sizes from an aerial perspective. Specifically, we observe that a substantial 47\% of instances are under 10 pixels, indicative of individuals who appear extremely small due to the altitude of the imagery. This reflects realistic scenarios where people are often tiny and challenging to detect.
The dataset also contains 52\% of instances in the range of 10-50 pixels and a minimal 1\% within the 50-300 pixel bracket. There are no instances above 300 pixels, reinforcing the dataset's focus on detecting smaller objects. In Table \ref{tab:instance_size_distribution}, when compared to datasets like PASCAL VOC and DOTA , the C2A dataset demonstrates a more balanced distribution between small and middle-sized instances. It is challenging for the models to detect the objects that are in tiny size. 
\vspace{-.6cm}
\begin{table}[t]
\centering
\caption{Comparison of instance size (in terms of width) distribution of some datasets in aerial images and natural images. Some statistics collected from \cite{xia2018dota}}
\label{tab:instance_size_distribution}
\begin{tabular}{@{}|l|c|c|c|c|@{}}
\toprule
Dataset               & <10 pixel & 10-50 pixel & 50-300 pixel & >300 pixel \\ \midrule
SARD                  & 0.01      & 0.66       & 0.32        & 0.01       \\
M4SFWD                & 0.02      & 0.38       & 0.51        & 0.09      \\
Smoke Scene           & 0      & 0.25       & 0.61        & 0.14      \\
Tiny Persons          & 0.56     & 0.42       & 0.02         & 0       \\
Crowd Human           & 0.01      & 0.42       & 0.50        & 0.07       \\
PASCAL VOC            & 0         & 0.14        & 0.61         & 0.25       \\
MSCOCO                & 0         & 0.43        & 0.49         & 0.08       \\
NWPU VHR-10           & 0         & 0.15        & 0.83         & 0.02       \\
3K Munich Vehicle & 0         & 0.93        & 0.07         & 0          \\
DOTA                  & 0         & 0.57        & 0.41         & 0.02       \\
\textbf{C2A (proposed)}   & \textbf{0.47} & \textbf{0.52} & \textbf{0.01} & \textbf{0}  \\
\bottomrule
\end{tabular}
\end{table}

\subsection{Aspect Ratio of Objects}
\begin{figure}[tb]
\centering
    \begin{subfigure}{.45\textwidth}
        \includegraphics[width=1.0\linewidth]{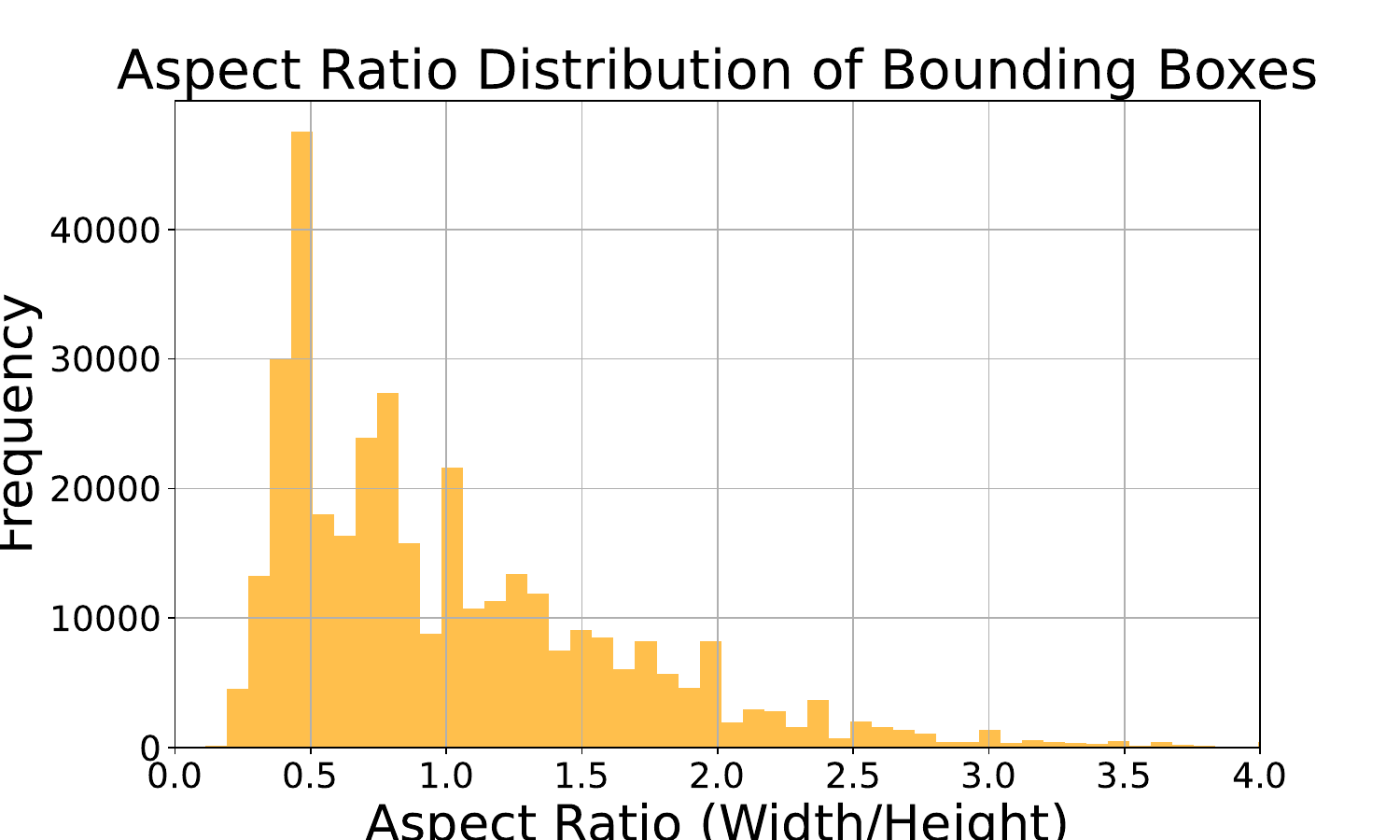}
        \caption{}
        \label{fig:aspect_ratio}
    \end{subfigure}
    \begin{subfigure}{0.45\textwidth}
        \includegraphics[width=1.0\linewidth]{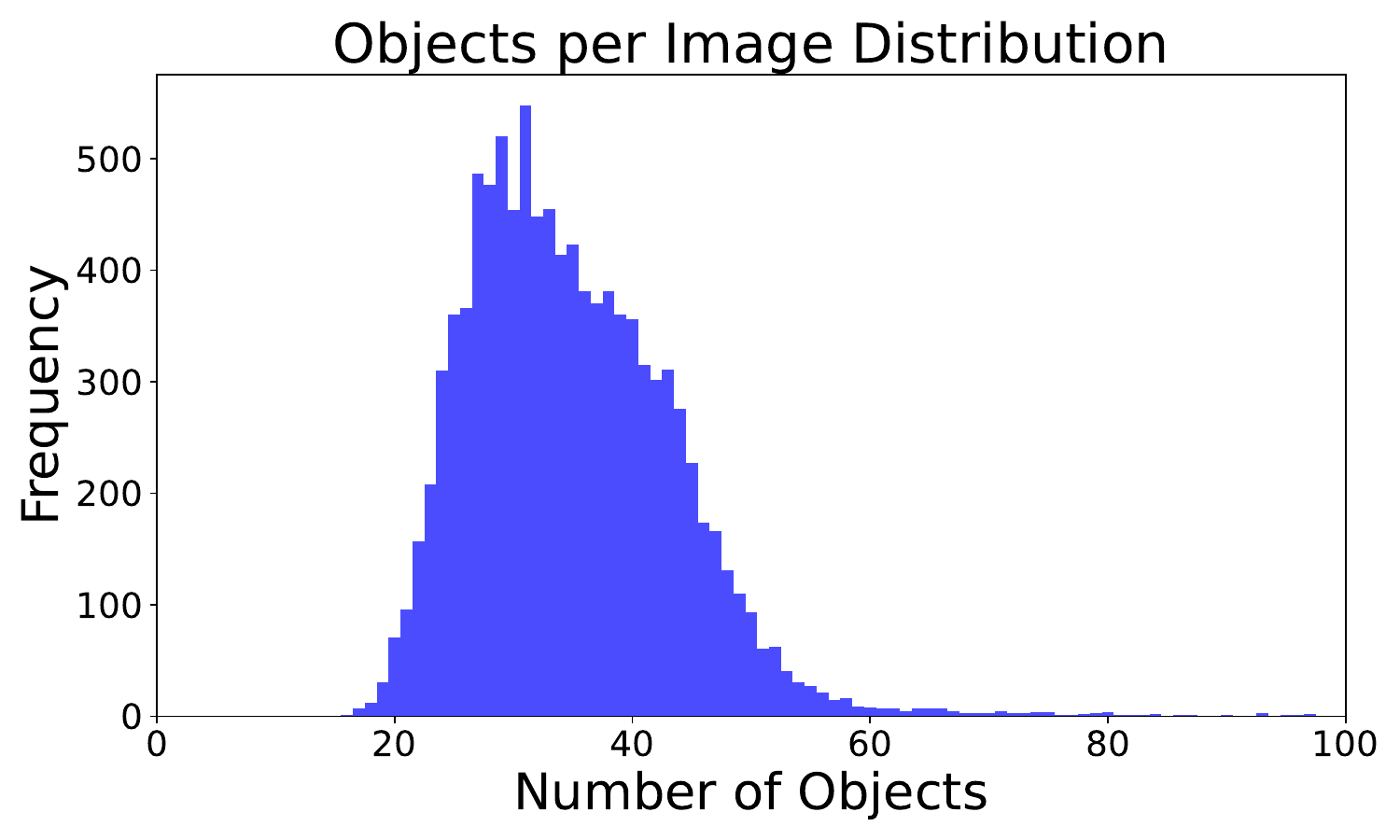}
        \caption{}
        \label{fig:object_density}
    \end{subfigure}
    \caption{(a) Aspect Ratio of C2A Dataset (b) Object Density}
\end{figure}

The aspect ratio (AR) is a critical parameter in anchor-based detection models, influencing the design and effectiveness of detectors like Faster R-CNN and YOLO series. In the C2A dataset, we analyze the AR of the minimally circumscribed horizontal bounding boxes encompassing each object. The histogram in Fig. \ref{fig:aspect_ratio} displays the distribution of these aspect ratios.
The distribution is skewed towards smaller ARs, with the majority of objects having an AR less than 1. This suggests that most bounding boxes are wider than they are tall, a likely scenario when dealing with collapsed individuals or those in horizontal positions in disaster scenarios. A noticeable amount of instances have ARs between 1 and 2, aligning with natural human proportions when standing or sitting. Very few instances possess a high AR, which is expected as elongated bounding boxes would be less common unless representing individuals in highly unusual orientations or in motion.


\subsection{Object Density of Images}
Aerial image datasets often exhibit a far greater number of objects per image when compared to datasets composed of natural images. Typical datasets, such as ImageNet, have an average of 2 objects per image, while MSCOCO averages 7.7. In stark contrast, our C2A dataset showcases a higher object density, reflective of the real-world complexity found in disaster-stricken environments.
\\
The histogram depicted in Fig. \ref{fig:object_density} outlines the frequency of object instances per image within our dataset. The distribution peaks significantly around 20 to 40 objects, with a notable extension towards images containing up to 100 instances. This dense distribution is a testament to the C2A dataset's capacity for providing a challenging and enriched learning context for object detection algorithms, pushing the envelope of their detection and discrimination capabilities.
\subsection{Human Pose and Disaster Scene Information}
In the pursuit of advancing SAR operations through machine learning, our C2A dataset offers more than object detection; it integrates critical contextual data by providing detailed annotations for both human poses and disaster scene types. The dataset categorizes human figures into five distinct poses: `Bent', `Kneeling', `Lying', `Sitting', and `Upright'. These annotations are crucial as they potentially correlate with the urgency and type of assistance required; for example, individuals found `Lying' or `Bent' in a disaster scene could indicate severe injury, necessitating immediate medical attention.
\\
Alongside pose information, the C2A dataset is annotated with disaster scene context, such as `traffic incident', `fire', `flood', and `collapsed building'. This level of detail allows for a nuanced understanding of the environment, providing vital clues about the challenges present in each unique scenario. Such information is instrumental for developing machine learning models that can not only detect humans in aerial images but also assess the severity and nature of the disaster context. The addition of these rich contextual layer opens new avenues for machine learning applications, potentially transforming the landscape of disaster response and emergency aid.

\section{Evaluation}
\label{sect4}
\subsection{Evaluation Metrics}

The evaluation of object detection models was conducted using the mean Average Precision (mAP) \cite{padilla2020survey}, a prevalent metric that integrates both precision and recall aspects of the predictions. Precision, defined as $\text{Precision} = \frac{TP}{TP + FP}$, measures the correctness of the predictions, while recall quantifies the model's ability to identify all relevant instances. The mAP is the mean of Average Precision (AP) across all classes, computed for varying Intersection over Union (IoU) thresholds, typically ranging from 0.5 to 0.95. The AP at a specific IoU threshold is the area under the precision-recall curve. The mAP at IoU threshold of 0.5, denoted as mAP@.50, is represented as $\text{mAP@.50} = \frac{1}{N} \sum_{i=1}^{N} \text{AP}_{i} \big|_{IoU=0.5}$, highlighting a model's proficiency in detecting objects with a moderate overlap with the ground truth.

\subsection{Training Options}
The evaluation of the models on the C2A dataset was conducted using NVIDIA A100 GPUs, with a uniform batch size of 24 and an image resolution of 640x640 pixels across 50 epochs. The ADAM optimizer was chosen for its efficiency in handling large datasets and complex image structures. Basic data augmentation techniques, such as flipping and resizing, were employed to enhance model robustness and prevent overfitting. The experiments were facilitated by popular deep learning frameworks, specifically mmDetection \cite{chen2019mmdetection}, Detectron2 \cite{wu2019detectron2}, and Ultralytics \cite{jocher2022ultralytics}, known for their high performance in object detection tasks. These frameworks provide extensive support for custom dataset training, enabling the effective application of state-of-the-art detection models to our specialized dataset.
\subsection{Benchmarking}
The C2A dataset was subjected to a rigorous evaluation process using a suite of state-of-the-art object detection models. These evaluations aimed to benchmark the dataset’s performance in training machine learning algorithms for the task of human detection in various disaster scenarios. The models were chosen for their relevance and proven accuracy in similar tasks, with an emphasis on assessing their capability to handle the complexities introduced by varied disaster backgrounds within the dataset. The Table \ref{tab:model_evaluation} illustrates the results found. 
\vspace{-.5cm}
\begin{table}[t]
\centering
\caption{Performance Evaluation of state-of-the-art Models on the C2A Dataset}
\label{tab:model_evaluation}
\begin{tabular}{|l|c|c||l|c|c|}
\hline
\textbf{Model} & \textbf{mAP} & \textbf{mAP@.50} & \textbf{Model} & \textbf{mAP} & \textbf{mAP@.50} \\ \hline
Faster R-CNN \cite{ren2015faster} & 0.3656 & 0.6340 & Dino \cite{liu2023grounding} & 0.4710 & 0.7890 \\ \hline
RetinaNet  \cite{lin2017focal}& 0.3834 & 0.6933 & Rtmdet \cite{lyu2022rtmdet}& 0.4420 & 0.7080 \\ \hline
Cascade R-CNN \cite{cai2018cascade}& 0.4860 & 0.7350 & YOLOv5 \cite{jocher2022ultralytics}& 0.4920 & 0.8080 \\ \hline
\textbf{YOLOv9-e} \cite{wang2024yolov9}& \textbf{0.6883} & \textbf{0.8927} &  YOLOv9-c \cite{wang2024yolov9}& 0.5562 & 0.7996  \\ \hline
\end{tabular}
\end{table}
\subsection{Result Analysis}
The evaluation results demonstrate a range of performance metrics across different models, reflecting the diverse strengths of each approach. YOLOv9-e outperformed other models with the highest mAP (mean Average Precision) score, indicating its superior ability to detect objects with a high degree of accuracy across varying Intersection over Union (IoU) thresholds. This suggests that the architectural improvements in YOLOv9, particularly for detecting small and partially occluded objects, are beneficial in the context of disaster scenarios.
\\
On the other hand, Faster R-CNN and RetinaNet, while offering competitive performance, particularly at the AP50 metric, fell short of the YOLO models. Dino and Cascade R-CNN showed substantial performance, with Cascade R-CNN achieving the second-highest mAP score, indicating its effectiveness in handling complex object relationships, likely due to its multi-stage detection process.
\\
The analysis of AP50 scores, which are based on a lower IoU threshold, reveals that most models perform significantly better when the requirement for the overlap between predicted and ground truth bounding boxes is relaxed. This discrepancy suggests that while the models are capable of identifying the presence of objects, refining the accuracy of bounding box predictions remains a challenge and an area for potential improvement in future research iterations.

\section{Discussion}
\label{sect5}
\subsection{Model Optimization for Complex Disaster Scenarios}
\label{model_opti}

\textcolor{black}{To investigate the impact of domain-specific training on model performance in complex disaster scenarios, we conducted a comparative analysis using several datasets and a model \cite{nihal2024blurry}: C2A (synthetic disaster scenes), SARD (real-world search and rescue images), and "General Human Detection" (a combination of crowd human \cite{shao2018crowdhuman}, tiny person \cite{yu2020scale}, and VisDrone \cite{du2019visdrone} datasets). By training models on these datasets and evaluating their performance across different validation sets, we aimed to identify the most effective approach for detecting humans in challenging disaster environments.
Table \ref{tab:comparative_performance} presents the results of this experiment, showcasing the mAP scores achieved by models trained on various datasets and validated on different test sets.}
\begin{table}[t]
\centering
\caption{\textcolor{black}{Comparative Performance Across Different Training and Validation Datasets}}
\label{tab:comparative_performance}
\begin{tabular}{|l|c|c|c|c|}
\hline
\textbf{Trained on\textbackslash Validated on} & \textbf{General Human} & \textbf{SARD} & \textbf{C2A} & \textbf{General Human+C2A} \\ \hline
General Human & 0.77 & 0.347 & 0.168 & 0.159 \\ \hline
SARD & 0.036 & 0.931 & 0.168 & 0.071 \\ \hline
C2A & 0.168 & 0.259 & 0.784 & 0.462 \\ \hline
General Human+C2A & 0.855 & 0.66 & 0.874 & 0.862 \\ \hline
\end{tabular}
\end{table}
\\\noindent\textcolor{black}{The model trained exclusively on the C2A dataset demonstrates a significant improvement in performance (0.784 mAP) when validated on the C2A test set compared to models trained on other datasets. This substantial increase in performance highlights the importance of domain-specific training using a dataset tailored to the task at hand, such as C2A, for developing models that can effectively detect humans in complex disaster scenarios.\\
One notable observation from the results is that although models trained on general human datasets perform poorly on search and rescue (SARD) and disaster (C2A) scenarios, the addition of the C2A dataset to the training process leads to a considerable improvement in performance. For instance, the model trained on the combined "General-Human + C2A" dataset achieves an mAP of 0.660 on the SARD validation set and 0.874 on the C2A validation set, surpassing the performance of models trained on either dataset alone. This finding suggests that incorporating disaster-specific data, such as the C2A dataset, can significantly enhance the model's ability to generalize to various challenging environments.\\
Furthermore, the results demonstrate that the combination of the C2A dataset and general human datasets yields better generalization performance across all validation sets. The model trained on the "General-Human + C2A" dataset achieves the highest mAP scores on the "General-Human" (0.855), SARD (0.660), and C2A (0.874) validation sets, indicating its robustness and versatility in handling diverse scenarios.\\
While the C2A dataset primarily consists of synthetic images, relying solely on synthetic data for training may undermine confidence in the model's real-world efficacy. Therefore, it is advisable to combine general human datasets with the C2A dataset to improve the model's ability to detect people in real-world disaster situations. The strong performance of the model trained on the "General-Human + C2A" dataset on the SARD validation set, which contains real-world search and rescue images, supports this recommendation.\\
To further validate the model's performance and increase confidence in its real-world applicability, future work should focus on evaluating the model on a more extensive set of real-world disaster images. This evaluation will help identify any potential gaps between the model's performance on synthetic and real-world data, guiding efforts to refine the dataset and training process.}

\subsection{Object Size and Detection Confidence}
\begin{figure}[tb]
\centering
    
    \begin{subfigure}{.45\columnwidth}
     \vspace*{.15cm}
        \includegraphics[width=1.0\linewidth]{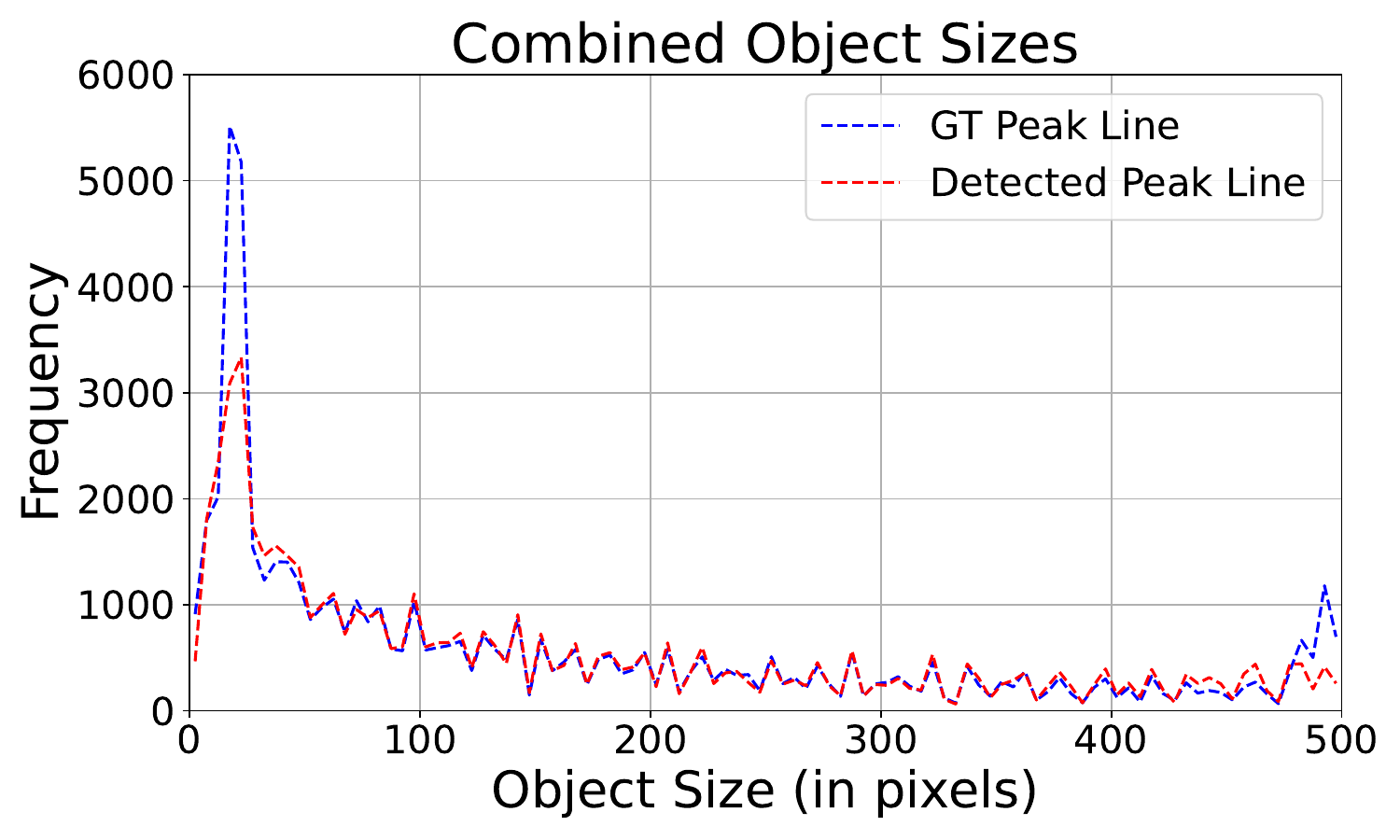}
        \hspace*{-3.5cm}\caption{}
    \end{subfigure}
    \begin{subfigure}{0.45\columnwidth}
        \includegraphics[width=1.05\linewidth]{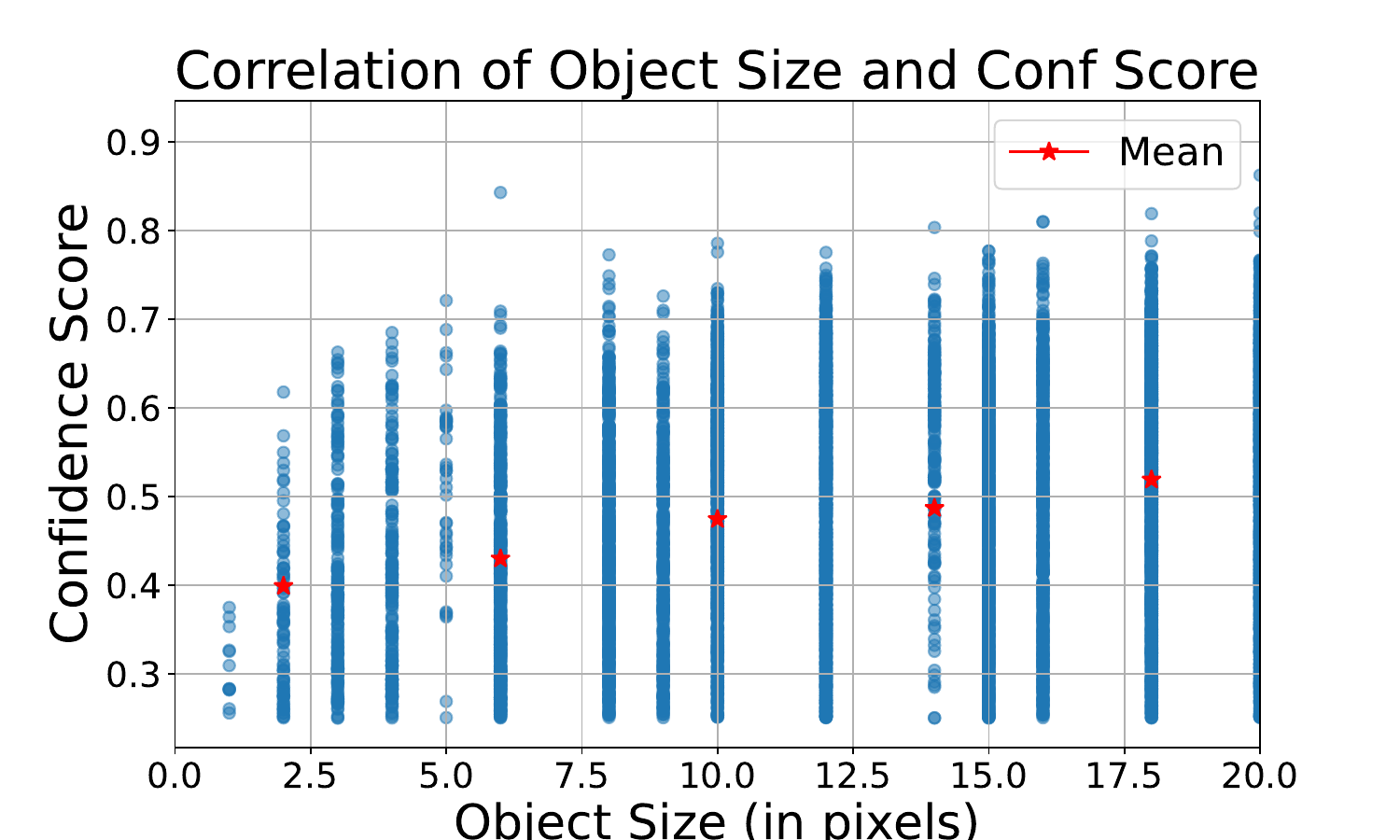}
        \caption{}
    \end{subfigure}
    \caption{\textcolor{black}{Comparative Analysis} of Object Detection (a) The frequency distribution of ground truth object sizes \textcolor{blue}{(blue)} showcases a clear decline in detection rates for smaller objects \textcolor{red}{(red)}, highlighting the challenges current detection algorithms face with objects less than 20 pixels in size. (b) The detection confidence scores across varying object sizes, with mean confidence indicated by red points, emphasize the higher reliability of detecting larger objects. These visualizations underscore the need for refining detection algorithms to better recognize small objects, which are critical for comprehensive disaster scene analysis.}
    \label{sp_comp}
\end{figure}

In-depth analysis of detection performance reveals a notable size bias where smaller objects (less than 20 pixels) are detected with less frequency and lower confidence scores. This trend, observable in Fig. \ref{sp_comp}(a), points to a potential size-dependent limitation inherent in current detection algorithms. Conversely, larger objects demonstrate higher detection confidence, as seen in Fig. \ref{sp_comp}(b), where the mean confidence score, represented by red points, scales with object size. This size-detection relationship suggests an avenue for model improvement—specifically, enhancing the sensitivity of detection algorithms to smaller objects could significantly improve performance in complex disaster environments, where small-scale features can be critical.

\subsection{Dataset Limitations and Prospects for Improvement}
The C2A dataset, while effective, encounters limitations due to its synthetic nature. The overlay of human figures from the LSP/MPII-MPHB dataset onto disaster scenes can sometimes result in unrealistic scaling and positioning, potentially compromising the model’s ability to generalize to real-world scenarios. Interestingly, this element of unrealism could also serve as a form of data augmentation, introducing variability that may help in training more robust and generalized models. Despite this, it is better to have context-aware adaptive scaling and improved spatial algorithms to enhance the realism of the training images. Moreover, transitioning to dynamic 3D models could more accurately depict human movement, overcoming the static nature of 2D images.  \textcolor{black}{Another limitation of the C2A dataset is that it consists of single images, whereas in most actual disaster scenarios, the input data could be in the form of video footage. This discrepancy between the training data and real-world application data may impact the model's performance in practical settings. Future work should focus on expanding the dataset to include video sequences of disaster scenes, enabling the development of models that can effectively process and analyze real-time video feeds from UAVs during SAR operations.} The dataset's variety in human poses and disaster scenarios is designed to aid in assessing disaster severity, enhancing its utility for SAR operations. Future enhancements should include real disaster footage to further validate and refine the dataset, optimizing AI model performance for real-world applications.

\section{Conclusion}
\label{sect6}

In the rapidly changing field of disaster response, our research introduces the C2A dataset as a crucial resource, connecting AI with humanitarian efforts. This work advances the technical capabilities of UAV-assisted search and rescue operations and represents a significant shift in how we integrate machine learning into crisis management. 
\\
The C2A dataset fills a critical need in disaster response, offering a comprehensive, synthetic environment that represents the complexities of real-world catastrophes. This dataset forms a foundation for training more robust and adaptable AI models. Our comparative analysis across various datasets highlights the importance of combining domain-specific data (like C2A) with general human datasets, resulting in models that are both specialized and widely applicable.
\\
These advancements are initial steps in an ongoing process. As we expand the capabilities of AI-assisted disaster response, we must consider the ethical implications and real-world applicability of our work. The limitations we've identified, particularly in synthetic data generation and real-world validation, serve as guides for future research.
\\
We envision a future where AI becomes an essential tool in crisis management, working alongside human expertise to save lives and reduce suffering. To achieve this vision, we encourage the research community to:
\begin{enumerate}
    \item Test the C2A dataset in real-world pilot studies, linking synthetic training with practical application.
    \item Collaborate to expand and refine the dataset, including diverse disaster scenarios and cultural contexts.
    \item Pursue research that combines computer vision, disaster management, and ethics to ensure responsible and effective use of AI in humanitarian efforts.
\end{enumerate}
As the field advances, our ultimate goal goes beyond technological progress; we aim to create tools and methods that are reliable in critical situations. By continually improving our approach, incorporating real-world feedback, and fostering collaboration across disciplines, we move towards a future where technology and human compassion work together, addressing the complexities of disaster response with increased precision and reliability.
\\
In conclusion, the C2A dataset and our findings represent not only a technical achievement but also progress towards a more resilient and responsive global community. As we confront increasingly complex global challenges, the combination of AI and human ingenuity offers hope for more effective, efficient, and compassionate disaster response strategies in the years to come.
\section*{Acknowledgements}
This work was supported by JSPS KAKENHI Grant No. JP22F22769 and JP22KF0141. Also, this work was performed the commissioned research fund provided by F-REI (JPFR23010102).
%
%
%
\bibliographystyle{splncs04}
\bibliography{refs}
%




\end{document}